\documentclass[letterpaper, 10 pt, conference]{ieeeconf} 

\IEEEoverridecommandlockouts                              % This command is only needed if 
\usepackage{cite}
\usepackage{times}
\usepackage{epsfig}
\usepackage{graphicx}
\usepackage{amsmath}
\usepackage{amssymb}
\usepackage{booktabs}

\usepackage{xcolor}
\usepackage{adjustbox}
\usepackage{multirow}
\usepackage{caption}
\usepackage{ tabularx}
\usepackage{url}
\usepackage{gensymb}

\usepackage{xcolor}

\usepackage[ruled,vlined]{algorithm2e}

\usepackage{lineno}
\modulolinenumbers[5]

\usepackage[pagebackref=true,breaklinks=true,letterpaper=true,colorlinks,bookmarks=false]{hyperref}

\let\labelindent\relax
\usepackage{enumitem}

\makeatletter
\newcommand{\printfnsymbol}[1]{%
  \textsuperscript{\@fnsymbol{#1}}%
}
\makeatother

\title{\LARGE \bf
Bidirectional Attention Network for Monocular Depth Estimation
}

\author{Shubhra~Aich$^*$,~Jean~Marie~Uwabeza~Vianney$^*$,~Md~Amirul~Islam,~Mannat~Kaur,~and~Bingbing~Liu$^\dagger$% <-this % stops a space
\thanks{\hspace{-2.5ex}$^*$Equal contribution; $^\dagger$corresponding author}
\thanks{\hspace{-2.5ex}Authors are with Noah's Ark Laboratory, Huawei Technologies, Markham, ON L3R5Y1, Canada.}
\thanks{\hspace{-2.5ex}\small Correspondence: liu.bingbing@huawei.com} 
}

\begin{document}

\maketitle
\thispagestyle{empty}
\pagestyle{empty}

%%%%%%%%%%%%%%%%%%%%%%%%%%%%%%%%%%%%%%%%%%%%%%%%%%%%%%%%%%%%%%%%%%%%%%%%%%%%%%%%
\begin{abstract}
In this paper, we propose a Bidirectional Attention Network (BANet), an end-to-end framework for monocular depth estimation (MDE) that addresses the limitation of effectively integrating local and global information in convolutional neural networks. The structure of this mechanism derives from a strong conceptual foundation of neural machine translation, and presents a light-weight mechanism for adaptive control of computation similar to the dynamic nature of recurrent neural networks. We introduce bidirectional attention modules that utilize the feed-forward feature maps and incorporate the global context to filter out ambiguity.  Extensive experiments reveal the high degree of capability of this bidirectional attention model over feed-forward baselines and other state-of-the-art methods for monocular depth estimation on two challenging datasets -- KITTI and DIODE. We show that our proposed approach either outperforms or performs at least on a par with the state-of-the-art monocular depth estimation methods with less memory and computational complexity.
\end{abstract}

%%%%%%%%%%%%%%%%%%%%%%%%%%%%%%%%%%%%%%
\section{Introduction}
Depth perception in human brain requires combined signals from three different cues (oculomotor, monocular, and binocular \cite{sensation-and-perception-goldstein}) to obtain ample information for depth sensing and navigation. As a matter of fact, the domain of MDE is inherently ill-posed. Classical computer vision approaches employ multi-view stereo correspondence algorithms~\cite{taxonomy-szeliski, compare-szeliski} for depth estimation.
Recent deep learning based methods formulate MDE as a dense, pixel-level, continuous regression problem \cite{densedepth, bts}, although a few approaches pose it as a classification \cite{mde-as-classification} or quantized regression \cite{dorn} task.

%Three different kinds of cues or signals are used for depth perception in the human brain: oculomotor, monocular, and binocular \cite{sensation-and-perception-goldstein}. The combination of all these cues provide ample information for depth sensing and navigation. As a matter of fact, the domain of monocular depth estimation (MDE) is inherently ill-posed.

State-of-the-art MDE models \cite{dorn, densedepth, bts} are based on pretrained convolutional backbones
\cite{vgg, resnet, densenet} with upsampling and skip connections \cite{densedepth}, global context module and
logarithmic discretization for ordinal regression \cite{dorn}, and coefficient learner for upsampling with local planar
assumption \cite{bts}. All these design choices explicitly or implicitly suffer from the spatial downsampling operation
in the backbone architecture which is shown for pixel-level task~\cite{amirul2017gated}. We incorporate the idea of depth-to-space (D2S) \cite{d2s, d2s-aich} transformation as a remedy to the downsampling operation in the decoding phase.
\begin{figure}[t]
%\vspace{-0.9cm}
	\begin{center}
		\includegraphics[width=0.49\textwidth]{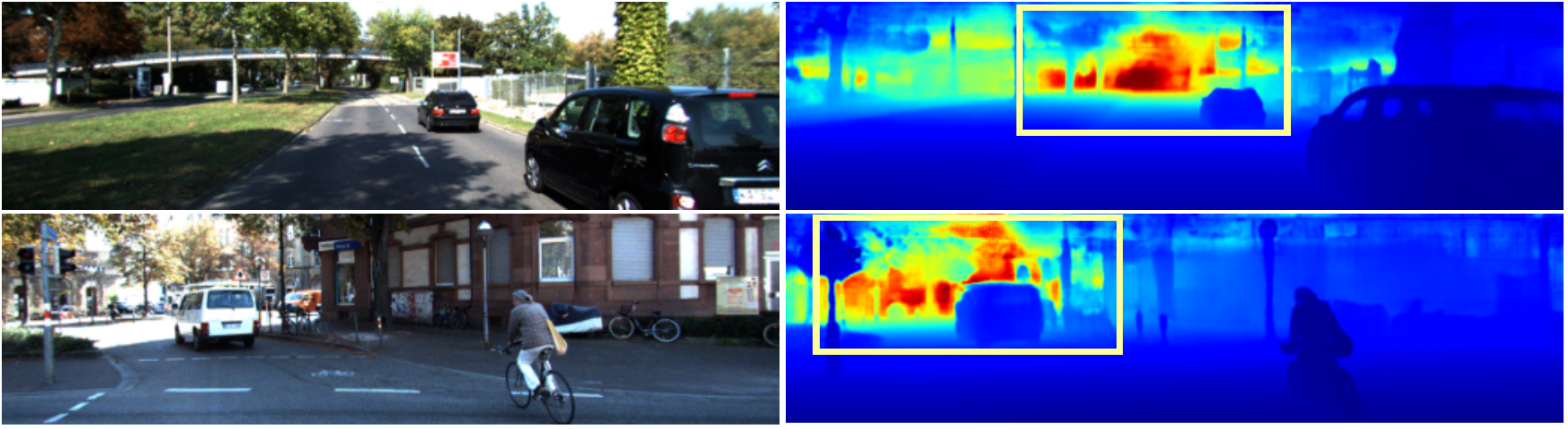}
		\vspace{-0.5cm}
		\caption{Sample BANet predictions on KITTI \textit{val} set. BANet improves the overall depth estimation by generating attention weights for each stage and diminishing representational ambiguity within the network.}
		\label{fig:intro}
	\end{center}
	\vspace{-1cm}
\end{figure}
However, straightforward D2S transformation of the final feature map might suffer from the lack of global context from the scene necessary
for reliable estimation \cite{how-see-depth, visualize-mde}. Therefore, we inject global context on the D2S transformed single-channel
feature maps for each stage of the backbone. 

\begin{figure*}[t!]
\centering
\adjustbox{max width=\textwidth}{
%\hskip-4.0cm
\includegraphics[width=0.75\textwidth]{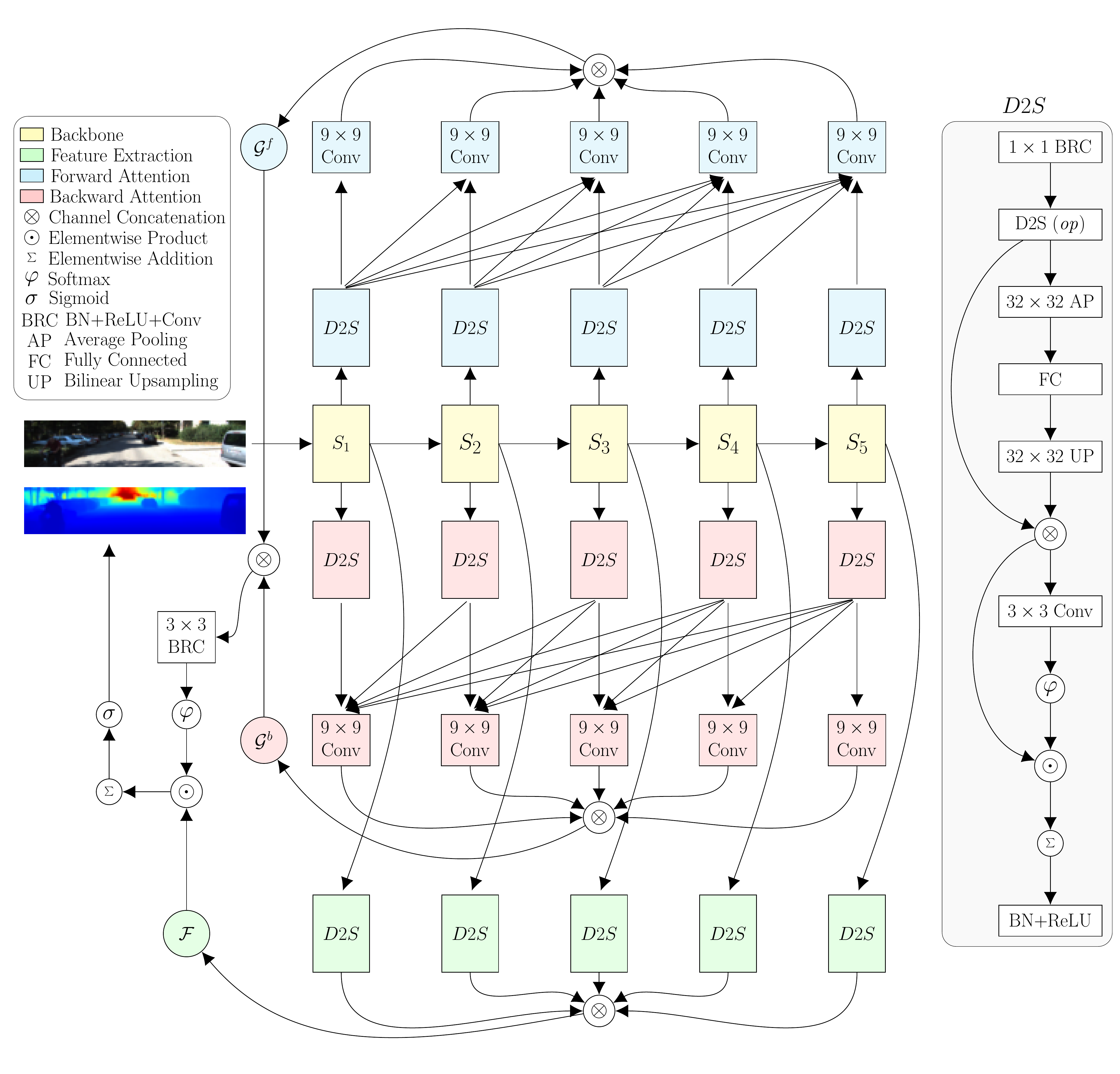}
} %
\vspace{-3ex}\caption{The computation graph of our proposed bidirectional attention network (BANet). The directed arrows refer to the flow of data through this graph. The forward and backward attention modules take stage-wise features as input and process through multiple operations to generate stage-wise attention weights. The rectangles represent either a single or composite operation with learnable parameters. The white circles with inscribed symbols ($\otimes, \varphi, \odot, \sum, \sigma$) denote parameterless operations. The colored circles highlight the outputs of different conceptual stages of processing. The construction of the $D2S$ modules is shown inside the gray box on the right. Note that the green $D2S$ modules used for feature computation $\mathcal{F}$ does not employ the last BN+ReLU block inside $D2S$.}
\label{fig:arch}
\vspace{-3ex}
\end{figure*}

Moreover, we effectively gather information from all the stages of the backbone with a bidirectional attention module (see Fig.~\ref{fig:arch}) inspired by \cite{nmt-attention} which first demonstrates the potential of attention in deep learning. Recently proposed attention-based works~\cite{senet, cbam, encnet, ecanet, attention-aug} mostly exploit either channel-wise or spatial attention in some form.

These approaches manipulate the internal feature maps of the backbone architecture with higher depth dimensions. For a RGB image with resolution, $H\times W \times 3$, the attention mechanism is applied on $\frac{H}{s} \times \frac{W}{s} \times D$ with stride, $s \in \{2,4,8,16,32\}$ and much higher depth D (generally between 256 and 2208). We hypothesize that the key aspect of a more accurate MDE architecture is its ability to produce the output with the same spatial resolution as input. Therefore, in our network design, we first transform all the backbone feature maps of dimension, $\frac{H}{s} \times \frac{W}{s} \times D$ into the original resolution of $H \times W \times 1$ followed by refinement process using our proposed bidirectional attention mechanism. Unlike other attention based approaches~\cite{senet, cbam, encnet, ecanet, attention-aug}, our bidirectional refinement process is applied outside the backbone with the original spatial resolution. This is efficiently doable for MDE as it is a single plane estimation task unlike the multi-class segmentation problem. Moreover, running the bidirectional attention modules on top of the single channel feature map brings additional advantages including computational efficiency and lower number of parameters. %All in all, the bidirectional attention mechanism is an efficient outer attention mechanism compared to the inner variants described in the recent computer vision literature. 
%This mechanism works perfectly by preserving the original spatial resolution with lower computational cost and memory and so with a reduced chance of overfitting.

Thus, in the light of the highlighted issues that arise in MDE methods, we present a novel yet effective pipeline for estimating continuous depth map from a single image (see Fig.~\ref{fig:intro}). Although our architecture contains substantially more connections than SOTA, both the computational complexity and the number of parameters are lower than the recent approaches as most of the interactions are computed on D2S transformed single channel features. Due to the prominence of bidirectional attention pattern in our design, we name our model \textit{Bidirectional Attention Network (BANet)}.

%We provide extensive experiments of several possible realizations of our BANet and other SOTA architectures on two large-scale MDE datasets: DIODE \cite{diode} and KITTI \cite{kitti}. We have chosen these two datasets due to both their scale and comparatively higher ground truth density than other smaller counterparts \cite{nyuv2, make3d-v3}. This is because the datasets with highly sparse ground truth \cite{nyuv2} are conventionally densified via interpolation, thus increasing the risk of assessment bias towards nonlinear interpolator than truly better depth estimator. 
\noindent \textbf{Contributions:} Our main contributions are as follows:
%Overall, our BANet performs better or on a par with the best SOTA architectures with both less parameters. Our main contributions are as follows:
\begin{itemize}[leftmargin=*]
    \item To the best of our knowledge, we present the first work applying the concept of bidirectional attention for monocular depth estimation. The flexibility of our mechanism is its ability to be incorporated with any existing CNNs.
    
    \item We further introduce \textit{forward} and \textit{backward} attention modules which effectively integrate local and global information to filter out ambiguity. 
    
    \item We present an extensive set of experimental results on two of the large scale MDE datasets to date. The experiments demonstrate the effectiveness of our proposed method in both the efficiency as well as performance. We show that variants of our proposed mechanism perform better or on a par with the SOTA architectures. 
    
\end{itemize}

\section{Related Work}
%Given the high degree of success of CNNs in image understanding, many methods~\cite{saxena-2006,mde-as-classification, structured-ms-crf,affinity,nrf,dorn,densedepth,lee-2019,deep-optics, mixing-dataset} have been proposed to tackle the problem of depth estimation. 

\noindent \textbf{Supervised MDE:}
A compiled survey of classical stereo correspondence algorithms for dense disparity estimation is provided in \cite{taxonomy-szeliski}. Saxena \textit{et al.} \cite{saxena-2006} employed Markov Random Field (MRF) to extract absolute and relative depth magnitudes using the local cues at multiple scales
to exploit both local and global information from image patches.
Based upon the assumption that the 3D scenes comprise many small planar surfaces (i.e. triangulation), this work is further extended to estimate the 3D position and orientation of the oversegmented superpixels in the images \cite{make3d-v1, make3d-v3}.

Eigen \textit{et al.} \cite{eigen} is the first work addressing the monocular depth estimation task
with deep learning. The authors proposed a stack of global, coarse depth prediction model followed by
local region-wise refinement with a separate sub-network.
Li \textit{et al.} \cite{li-crf} refined the superpixel depth map predicted by the CNN model
to the pixel-level using a CRF.
Cao \textit{et al.} \cite{mde-as-classification} transformed the continuous regression task of MDE to
the pixel-level classification problem by discretization of depth ranges into multiple bins. %The classification probability is further utilized to refine the predictions with a fully-connected CRF.
A unified approach blending CRF and fully convolutional networks with superpixel pooling for faster
inference is proposed into the framework of Deep Convolutional Neural Fields (DCNF) \cite{dcnf}.
Multi-scale outputs from different stages of CNN are fused with continuous CRFs in \cite{ms-crf} with
further refinements via structured attention over the feature maps \cite{structured-ms-crf}.

%Joint prediction of related attributes, such as depth, surface normal, and semantic labels is proposed in \cite{eigen-2015, wang-2015}. Laina \textit{et al.} \cite{laina-2016} proposed a fully convolutional network equipped with residual learning and a faster upsampling based on unpooling to improve the resolution of depth prediction. The neural regression forest (NRF) \cite{nrf} attempted to utilize the expressive power of CNNs and the efficacy of regression forests for small datasets using a binary regression tree with shallow CNNs as its nodes. %Despite quite shallow models at each node, NRF as a whole is equivalent to a deep CNN.

%Affinity among the neighboring pixels are modeled with the similarity of pixel features in \cite{affinity}. Also, vertical (unidirectional) pooling is applied to capture the drastic changes in depth that occurs in a real world scenario. Yin \textit{et al.} \cite{virtual-normal} harness the geometric constraint with the loss between the virtual normal computed from the set of three 3D points in predicted and ground truth depth maps. Besides absolute depth, relative depth or the ratio of depth values between pairs of sub-regions in the image is used for optimal depth prediction in \cite{lee-2019}. Deep Optics \cite{deep-optics} employs an optical-encoder and electronic-decoder system to exploit the advantages of freeform lenses over conventional ones.

Recently, DORN \cite{dorn} modeled the MDE task as an ordinal regression problem with space-increasing (logarithmic)
discretization of the depth range to appropriately address the increase in error with depth magnitude.
DenseDepth \cite{densedepth} employed the pretrained DenseNet \cite{densenet} backbone with bilinear
upsampling and skip connections on the decoder to obtain high-resolution depth map.
The novelty in the BTS architecture \cite{bts} lies in their local planar guidance (LPG) module as an
alternative to upsampling based skip connections to transform intermediate feature maps to full-resolution depth
predictions. In contrast, we propose a light-weight bidirectional attention mechanism to filter out ambiguity from the deeper representations by incorporating global and local contexts. 

\noindent \textbf{Visualization of MDE models}:
%To date, there exist few works \cite{visualize-mde,how-see-depth} on the visualization and possible interpretation of the deep learning models for monocular depth estimation. 
Hu \textit{et al.} \cite{visualize-mde} showed that CNNs most importantly need the edge pixels to
infer the scene geometry and regions surrounding the vanishing points for outdoor scenarios.
Dijk and Croon \cite{how-see-depth} found that the pretrained MDE models ignore the apparent size of the objects and focus only on the vertical position of the object in the scene to predict the distance.
It means that the models approximate the camera pose while training. This is particularly
true for the KITTI dataset for which the camera is mounted on top of the vehicle and the vanishing point of all the scenes is around the middle of the images. Consequently, a small change in pitch corrupts the depth prediction in a non-negligible manner. 

\section{Our Approach}
In this section, we discuss our proposed Bidirectional Attention Network (BANet). Specifically, we introduce  bidirectional attention modules  and  different global context aggregation techniques for effective integration of local and global information for the task of monocular depth estimation.

\subsection{Bidirectional Attention Network}

The idea of applying bidirectional attention in our proposed approach is motivated by neural machine translation (NMT)~\cite{nmt-attention}. Although there exist recent works~\cite{senet, cbam, encnet, ecanet, attention-aug} that exploits channel-wise and spatial attention in CNN for various computer vision tasks, the idea of applying attention in forward and backward manner to achieve the nature of birectional RNN has not been explored yet.    

%The architecture proposed in this paper is a convolutional realization of the phenomenal paper regarding the use of attention in a bidirectional RNN model for neural machine translation (NMT) \cite{nmt-attention}. Although there are several influential papers in the domain of computer vision exploiting channel-wise and spatial attention in CNN in recent years \cite{senet, cbam, encnet}, there is not much effort to emulate the idea of bidirectional attention similar to RNN for vision tasks.

%Our overall architecture is illustrated in Fig.~\ref{fig:arch}. In general, in all the standard backbones for image classification, such as AlexNet \cite{alexnet}, VGG \cite{vgg}, Inception \cite{inception-v3}, ResNet \cite{resnet}, DenseNet \cite{densenet}, SENet \cite{senet}, etc., all the computations are arranged into five feedforward stages. The input to the first stage is the RGB image, the input to the second stage is the output of the first stage, and so on. Although SOTA architectures like ResNet or DenseNet employ residual and dense connections between different computational blocks, these connections exist only within the blocks of the same stage. That said, there is no residual or dense connections between two different blocks residing in two different stages. The notion of this stage is more computational where the filter growth rate (DenseNet) or the number of filters (ResNet, VGG, etc.) and the spatial resolution are mostly the same inside each stage of computation. 

Our overall architecture is illustrated in Fig.~\ref{fig:arch}. In general, standard CNNs for image classification \cite{alexnet, vgg, inception-v3, resnet, densenet, senet} consists of five feed-forward stages. We denote the final feature map from stage $i$ as $s_i$, thus resulting in an ordered set of five feature maps, $\{s_1, s_2, s_3, s_4, s_5\}$. 
To draw an analogy with the words in a sentence used in NMT \cite{nmt-attention}, this set can be portrayed as a visual sentence (ignoring their input-output dependencies) where each $s_i$ is analogous to a word. Note that the bidirectional RNN inherently generates the forward and backward hidden states due to the sequential and dynamic processing of the source sentence word by word. Since CNNs have a static nature on the input images, we introduce forward and backward attention sub-modules which take the stage-wise feature maps, $s_i$ as input.
\vspace{0.2ex}\\
\noindent \textbf{Bidirectional Attention Modules:} First, we reshape the stage-wise feature maps, $s_i$, (where $i=1,2,..,5$) to the output spatial resolution using a $1\times 1$ convolution layer followed by the depth-to-space (D2S) \cite{d2s,d2s-aich} operation. Our forward and backward attentions are given by:

\vspace{-2ex}
% \begin{equation}
% \begin{split}
% &s^f_i = \mathcal{D}^f_i \left ( s_i \right ) \quad \quad \quad \quad \quad \quad \quad \quad \;\: \forall i \in \{1,2, \hdots, N\} \\
% &a^f_i = \mathcal{A}^f_i \left ( s^f_1, s^f_2, \hdots, s^f_i \right ) \quad \quad \quad \forall i \in \{1,2, \hdots, N\} \\
% &\mathcal{G}^f = a^f_1 \otimes a^f_2 \otimes \hdots \otimes a^f_N \\
% &s^b_i = \mathcal{D}^b_i \left ( s_i \right ) \quad \quad \quad \quad \quad \quad \quad \quad \;\;\; \forall i \in \{1,2, \hdots, N\} \\
% &a^b_i = \mathcal{A}^b_i \left ( s^b_i, s^b_{i+1}, \hdots, s^b_N \right ) \quad \quad \;\;\; \forall i \in \{1,2, \hdots, N\} \\
% &\mathcal{G}^b = a^b_1 \otimes a^b_2 \otimes \hdots \otimes a^b_N \\
% &\mathcal{A} = \varphi \left ( \mathcal{G} \left ( \mathcal{G}^f \otimes \mathcal{G}^b \right ) \right ) \\
% \end{split}
% \label{eq:fb_attention}
% \vspace{-3ex}
% \end{equation}

\vspace{0cm}
\begin{equation}
\centering
\begin{split}
&s^f_i = \mathcal{D}^f_i \left ( s_i \right ); \hspace{2.3cm} 
s^b_i = \mathcal{D}^b_i \left ( s_i \right )\\
&a^f_i = \mathcal{A}^f_i \left ( s^f_1, s^f_2, \hdots, s^f_i \right );  \hspace{0.45cm}
a^b_i = \mathcal{A}^b_i \left ( s^b_i, s^b_{i+1}, \hdots, s^b_N \right )  \\
&\mathcal{G}^f = a^f_1 \otimes a^f_2 \otimes \hdots \otimes a^f_N; \hspace{0.5cm}
\mathcal{G}^b = a^b_1 \otimes a^b_2 \otimes \hdots \otimes a^b_N \\
&\mathcal{A} = \varphi \left ( \mathcal{G} \left ( \mathcal{G}^f \otimes \mathcal{G}^b \right ) \right )
\end{split}
\label{eq:fb_attention}
\vspace{-4ex}
\end{equation}

\begin{figure*}[t!]
%\vspace{-0.9cm}
	\begin{center}
		\includegraphics[width=0.99\textwidth]{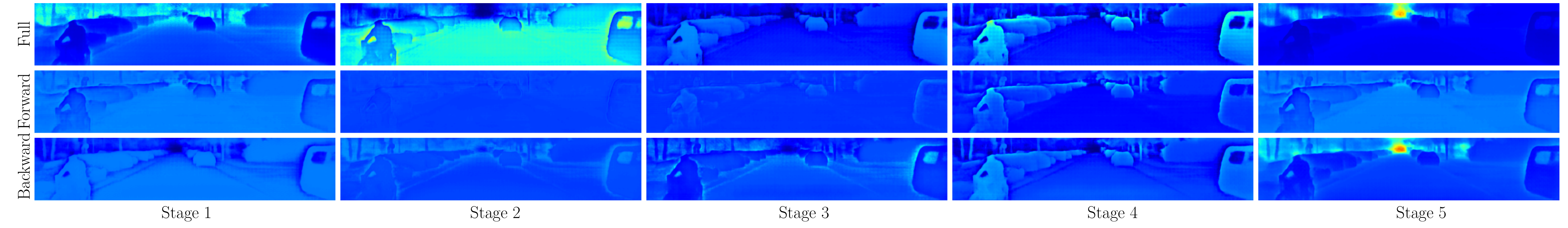}
		\vspace{-0.2cm}
		\caption{An illustration of the generated stage-wise attention weights using the forward and backward attention modules.}
		\label{fig:pseudo}
	\end{center}
	\vspace{-0.8cm}
\end{figure*}

$\forall i \in \{1,2, \hdots, N\}$. In Eq.~\ref{eq:fb_attention}, the superscripts $\square^f$ and $\square^b$ denote the operations related to the forward and backward attentions, respectively, and the subscript $\square_i$ denotes the associated stage of the backbone feature maps. 
$\mathcal{D}^f_i$ and $\mathcal{D}^b_i$ represent the $D2S$ modules which process the backbone feature map $s_i$. $\mathcal{A}^f_i$ and $\mathcal{A}^b_i$ refer to the $9\times 9$ convolution in Fig.~\ref{fig:arch}. Note that $\mathcal{A}^f_i$ has access to the features up to the $i^{th}$ stage, and $\mathcal{A}^b_i$ receives the feature representation from the $i^{th}$ stage onward; thus emulating the forward and backward attention mechanisms of a bidirectional RNN. Next, all the forward and backward attention maps are concatenated channel-wise and processed with a $3\times 3$ convolution ($\mathcal{G}$ in Eq.~\ref{eq:fb_attention}) and softmax ($\varphi$) to generate the per stage pixel-level attention weights, $\mathcal{A}$ (see Fig.~\ref{fig:pseudo}). The procedure of computing the feature representation, $f_i$, from the stage-wise feature maps, $S_i$, using the D2S module can be formalized as: 
\vspace{-1ex}
\begin{equation}
\begin{split}
&f_i = \mathcal{D}^\mathcal{F}_i \left ( s_i \right ); \hspace{0.4cm} \mathcal{F} = f_1 \otimes f_2 \otimes \hdots \otimes f_N \\
\end{split}
\label{eq:fb_feature}
\vspace{-1ex}
\end{equation}

$\forall i \in \{1,2, \hdots, N\}$. Then we compute the unnormalized prediction, $\widehat{D}_u$, from the concatenated features, $\mathcal{F}$, and attention maps, $\mathcal{A}$, using the Hadamard (element-wise) product followed by pixel-wise summation $\sum$. Finally, the normalized prediction, $\widehat{D}$, is generated using the sigmoid ($\sigma$) function. The operations can be expressed as:
\vspace{0ex}
\begin{equation}
\begin{split}
&\widehat{D}_u = \sum  \mathcal{A} \odot \mathcal{F}, \hspace{0.4cm} \widehat{D} = \sigma \left ( \widehat{D}_u \right ) \\
\end{split}
\label{eq:fb_feature}
\vspace{-1ex}
\end{equation}
%%%%%%%%%%%%%%%

\noindent \textbf{Global Context Aggregation.}
%Our Global-Local Interaction (GLI) module is an efficient realization of the attention [] mechanism.
We incorporate global context inside the $D2S$ modules by applying average pooling with relatively large kernels followed by fully connected layers and bilinear upsampling operations. 
%This set of operations combine the pixel-wise, local (query) information with the image-level, global (key) information to extract better monocular cues from the whole image. 
This aggregation of global context in the D2S module helps to resolve ambiguities (see Fig.~\ref{fig:diode_results} and \ref{fig:kitti_results}) for thinner objects in more challenging scenarios (i.e., very bright or dark contexts). 
%More visualizations demonstrating the effect of global context aggregation can be found in the supplementary material.
%We do not find the attention factor of relative position [ ] assisting in any potential gain in performance. 
%At first, it seems somewhat counterintuitive to the recent work \cite{visualize-mde} showing the importance of the vanishing point in the outdoor scenarios for monocular depth estimation. One possible explanation to this problem might be the static nature of the available datasets. For example, for outdoor scenarios, if more or less visible, the vanishing point is located around the center of the image where the road merges with the sky at the horizon; featuring a highly discriminative set of spatial cues. Thus, the network can easily modulate a subset of its kernel weights to store this information based only upon the global (key) content. On the other hand, for indoor images, although the scene graph is less dynamic than outdoor ones, both the scene structure and the texture patterns are arguably more complex due to intricate details arising from the interior design and furnitures. Moreover, the location of the vanishing point varies substantially depending upon both the scene and the camera angle. Consequently, the narrower range of depth for indoor scenes makes the local and semi-local features more important than the vanishing point based global cues.
Additionally, we provide details of few alternative realizations of our proposed architecture as follows:
\begin{itemize}[leftmargin=*]
	\item \textbf{BANet-Full:} This is the complete realization of the architecture as shown in Fig.~\ref{fig:arch}.
	\item \textbf{BANet-Vanilla:} It contains only the backbone followed by $1\times 1$ convolution, a single D2S operation, and sigmoid to generate the final depth prediction. This is similar to \cite{refinedmpl}.
	\item \textbf{BANet-Forward}: The backward attention modules of BANet are missing in this setup.
	\item \textbf{BANet-Backward:} The forward attention modules of BANet are not included here.
	\item \textbf{BANet-Markov:} This follows the Markovian assumption that the feature at each time step (or stage) $i$ depends only on the feature from the immediate preceding (for forward attention) or succeeding (for backward attention) time step (or stage) $i\mp 1$ (i.e. $X_i \perp \{ X_{i\mp 2}, X_{i\mp 3}, ... X_{i\mp k} \} | X_{i\mp 1} $). %Therefore, all but the immediate preceding (forward) and succeeding (backward) incoming edges to the $9\times 9$ convolution in Fig.~\ref{fig:arch} are deactivated in this construct.
	\item \textbf{BANet-Local:} This variant replaces the global context aggregation part with a single $9\times 9$ convolution.
	
\end{itemize}

To demonstrate that the performance improvement on the MDE task using the time dependent structuring is not due to the mere increase in the capacity as well as the number of parameters, we conduct pilot experiments without any time dependent structuring by simply concatenating the different stage features all at once followed by similar post-processing. However, we empirically find that such a naive implementation performs much poorer than our proposed time dependent realizations mentioned above. Therefore, we exclude this straightforward employment from further experimental analysis.

\vspace{-1ex}
\section{Experiments}
%To evaluate the effectiveness of our proposed bidirectional attention framework, we first conduct ablation study with different variants and provide quantitative comparison on two challenging monocular depth estimation dataset (KITTI and DIODE).  Next, we provide a comparison with recent state-of-the-art methods on KITTI test set.  

\noindent \textbf{Datasets:} Among several publicly available standard datasets \cite{nyuv2,diode,kitti,make3d-v1,make3d-v3} for monocular depth estimation, we choose DIODE \cite{diode} and KITTI \cite{kitti} due to their comparatively high density of ground truth annotations. 
%For example, in NYUv2 \cite{nyuv2} dataset, the sparse ground truth is densified with interpolation based on image inpainting. 
We hypothesize that our model assessment might be biased towards better nonlinear interpolator rather than the candidates capable of learning to predict depth maps using monocular cues. Therefore, our experiments done on the datasets comprising truly dense ground truths provide unbiased measure of different types of architectural design choices.\\
$\bullet$ \textbf{DIODE}: Dense Indoor/Outdoor DEpth (DIODE) \cite{diode} is the first standard dataset for monocular depth estimation comprising diverse indoor and outdoor scenes acquired with the same hardware setup. 
%FARO Focus S350 \cite{faro-scanner} is used as the laser scanner for data collection. 
%The minimum and maximum range of this sensor is $0.6m$ and $350m$, respectively. This range is comparatively much higher than the counterparts used in other datasets (i.e. the maximum range for KITTI is $120m$ with a Velodyne HDL-64E), which makes the prediction tasks considerably more challenging, esp. for the outdoor scenes. Moreover, both angular resolution ($0.009 \degree$) and depth precision ($\pm 1 mm$) for the sensor have significantly higher granularity leading to denser and more reliable ground truth for depth estimation.
The training set consists of 8574 indoor and 16884 outdoor samples from 20 scans each. The validation set contains 325 indoor and 446 outdoor samples with each set from 10 different scans. The ground truth density for the indoor training and validation splits are approximately 99.54$\%$  and 99$\%$, respectively. The density of the outdoor sets are naturally lower with 67.19$\%$ for training and 78.33$\%$ for validation subsets. The indoor and outdoor ranges for the dataset are $50m$ and $300m$, respectively.\\
$\bullet$ \textbf{KITTI}:
%KITTI \cite{kitti-org} dataset is the \textit{de facto} standard for evaluating the vision algorithms for autonomous driving. 
KITTI dataset for monocular depth estimation \cite{kitti} is a derivative of the KITTI sequences. 
%\textit{Velodyne HDL-64E} is used as the laser scanner for KITTI that also provides raw, semi-dense ground truth for depth prediction. The refined depth ground truth is generated by stereo reconstruction based post-processing \cite{kitti}. 
The training, validation, and test sets comprise 85898, 1000, and 500 samples, respectively. Following previous works, we set the prediction range to $[0-80m]$ where ever applicable, although there are a few pixels with depth values greater than $80m$.\\
\noindent \textbf{Implementation Details:}
For fair comparison, all the recent models \cite{dorn, bts, densedepth} are trained from scratch alongside ours using the hyperparameter settings proposed in the corresponding papers. Our models are trained for 50 epochs with batch size of 32 in the \texttt{NVIDIA Tesla V100 32GB GPU} using the sparse labels without any densification. The runtime was measured with a single \texttt{NVIDIA GeForce RTX 2080 Ti GPU} just because of its availability at the time of writing.
We use Adam optimizer with the initial learning rate of $1e^{-4}$ and weight decay of 0. The learning rate is multiplied on plateau after 10 epochs by 0.1 until $1e^{-5}$. Random horizontal flipping and color jitter are the only augmentations used in training. For KITTI, we randomly crop $352 \times 1216$ training samples following the resolution of the validation and test splits. Also, for computational efficiency, the inputs are half-sampled (and zero-padded if necessary) prior to feeding into the models, and the predictions are bilinearly upsampled to match the ground truth before comparison. For DORN~\cite{dorn}, we set the number of ordinal levels to the optimal value of 80 (or 160 considering both positive and negative planes). This value is set to 50, 75, and 75 for DIODE indoor, outdoor, and combined splits following the same principle. 
We employ SILog and $\mathcal{L}_1$ losses for training all the models on KITTI and DIODE, respectively, except for DORN that comes with its own ordinal loss. Following BTS~\cite{bts} and DenseDepth~\cite{densedepth}, we use DenseNet161~\cite{densenet} as our BANet backbone. \\
\noindent \textbf{Evaluation Metrics:}
%In MDE literature, both accuracy (higher is better) and error (lower is better) metrics are used altogether to benchmark different approaches. However, there is a lack of consistency among the set of metrics used for different datasets. In this work, we employ a unified set of metrics across all the subsets of different datasets used for experiments. 
For the error (lower is better) metrics, we mostly follow the measures (SILog, SqRel, AbsRel, MAE, RMSE, iRMSE) directly from the KITTI leaderboard. 
Additionally, we have used the following thresholded accuracy (higher is better) metrics (Eq. \ref{eq:acc-metrics}). 
\begin{equation}
\begin{split}
\vspace{-10ex}
%&\textbf{Extended accuracy metrics}\\
&\delta_k = \frac{1}{N} \sum_{i} { \left ( max(\frac{a_i}{t_i} \frac{t_i}{a_i}) < (1 + \frac{k}{100}) \right ) } \;\;\; * 100 \;\;(\%);\\ 
& \hspace{5.0cm}k \in \{ 25, 56, 95 \}
\end{split}
\label{eq:acc-metrics}
\end{equation}

\begin{table*}[]
\centering
\caption{Quantitative results on the \textbf{DIODE Indoor} \textit{val} set.}
\vspace{-2ex}
\label{tab:diode-indoor}
\adjustbox{max width=0.85\textwidth}{ %
%\hskip-4.0cm
\begin{tabular}{l|c|c|cccccc|ccc}
\specialrule{1.2pt}{1pt}{1pt}
\multicolumn{1}{c|}{\multirow{2}{*}{Model}} & \multirow{2}{*}{\begin{tabular}[c]{@{}c@{}}\#Params\\ ($10^6$)\end{tabular}} & \multirow{2}{*}{\begin{tabular}[c]{@{}l@{}}Time\\ (ms)\end{tabular}} & \multicolumn{6}{c|}{Lower is better} & \multicolumn{3}{c}{Higher is better (\%)} \\ \cline{4-12}
 &  &  & \textbf{SiLog} & \textbf{SqRel} & \textbf{AbsRel} & \textbf{MAE} & \textbf{RMSE} & \textbf{iRMSE} & $\delta_{25}$ & $\delta_{56}$ & $\delta_{95}$ \\ 
 \specialrule{1.2pt}{1pt}{1pt}
 
DORN \cite{dorn} & 91.21 & 57 & 24.45 & 49.54 & 47.25 & 1.63 & 1.86 & 208.30 & 37.61 & 62.39 & 77.97 \\
%DORN-BB \cite{dorn} & 91.01 & 50 &  &  &  &  &  &  &  &  &  &  &  &  \\
DenseDepth \cite{densedepth} & 44.61 & 27 & 19.67 & 25.10 & 34.50 & \textbf{1.39} & \textbf{1.56} & 177.33 & 48.13 & 70.19 & 83.00 \\
BTS \cite{bts} & 47.00 & 34 & 19.19 & 25.55 & 33.31 & 1.42 & 1.61 & \textbf{171.41} & 50.14 & \textbf{72.71} & \textbf{83.44} \\
\hline

BANet-Vanilla & 28.73 & 21 & 18.78 & 27.75 & 34.36 & 1.43 & 1.61 & 175.35 & \textbf{51.28} & 70.63 & 81.32 \\
BANet-Forward & 32.58 & 28 & 18.39 & 25.81 & 35.49 & 1.49 & 1.65 & 179.58 & 47.53 & 68.63 & 80.91 \\
BANet-Backward & 32.58 & 28 & \textbf{18.25} & 24.43 & 34.19 & 1.46 & 1.62 & 176.02 & 48.94 & 69.50 & 81.52 \\
BANet-Markov & 35.64 & 31 & 18.62 & \textbf{21.78} & \textbf{32.58} & 1.44 & 1.61 & 173.58 & 48.99 & 72.28 & 82.29 \\
BANet-Local & 35.08 & 32 & 18.83 & 23.65 & 33.49 & 1.43 & 1.59 & 172.11 & 49.71 & 71.18 & 82.88 \\
BANet-Full & 35.64 & 33 & 18.43 & 26.49 & 34.99 & 1.45 & 1.61 & 177.22 & 48.48 & 70.66 & 82.26 \\ \specialrule{1.2pt}{1pt}{1pt}
\end{tabular} } %
\vspace{-2.5ex}
\end{table*}
%%%%%%%%%%%%%%%%%%%%%%%%%%%%%%%%
\begin{table*}[]
\centering
\caption{Quantitative results on the \textbf{DIODE Outdoor} \textit{val} set.}
\vspace{-2ex}
\label{tab:diode-outdoor}
\adjustbox{max width=0.85\textwidth}{ %
%\hskip-4.0cm
\begin{tabular}{l|c|c|cccccc|ccc}
\specialrule{1.2pt}{1pt}{1pt}
\multicolumn{1}{c|}{\multirow{2}{*}{Model}} & \multirow{2}{*}{\begin{tabular}[c]{@{}c@{}}\#Params\\ ($10^6$)\end{tabular}} & \multirow{2}{*}{\begin{tabular}[c]{@{}l@{}}Time\\ (ms)\end{tabular}} & \multicolumn{6}{c|}{Lower is better} & \multicolumn{3}{c}{Higher is better (\%)} \\ \cline{4-12}
 &  &  & \textbf{SiLog} & \textbf{SqRel} & \textbf{AbsRel} & \textbf{MAE} & \textbf{RMSE} & \textbf{iRMSE} & $\delta_{25}$ & $\delta_{56}$ & $\delta_{95}$ \\ 
 \specialrule{1.2pt}{1pt}{1pt}

DORN \cite{dorn} & 91.31 & 57 & 40.96 & 249.72 & 48.42 & 5.17 & 8.06 & 98.01 & 46.92 & 75.36 & 87.30 \\
%DORN-BB \cite{dorn} & 91.01 & 50 &  &  &  &  &  &  &  &  &  &  &  &  \\
DenseDepth \cite{densedepth} & 44.61 & 27 & 37.49 & 244.47 & 41.74 & 4.32 & 7.03 & 110.19 & 58.56 & 81.14 & 90.05 \\
BTS \cite{bts} & 47.00 & 34 & 39.91 & 236.51 & 41.86 & 4.40 & 7.15 & 93.00 & 57.71 & 80.62 & 89.73 \\
\hline

BANet-Vanilla & 28.73 & 21 & 37.38 & 217.52 & 39.53 & 4.19 & 6.93 & 92.85 & 58.95 & 81.42 & 90.09 \\
BANet-Forward & 32.58 & 28 & 37.59 & \textbf{202.70} & 38.80 & 4.25 & 7.00 & 93.40 & 58.46 & 80.81 & 89.72 \\
BANet-Backward & 32.58 & 28 & 36.89 & 215.27 & 39.15 & 4.21 & 6.86 & 92.44 & 57.89 & 81.34 & 90.37 \\
BANet-Markov & 35.64 & 31 & 37.61 & 226.07 & 39.86 & 4.27 & 6.97 & 92.86 & 58.46 & 81.24 & 90.00 \\
BANet-Local & 35.08 & 32 & \textbf{36.54} & 218.54 & 39.23 & \textbf{4.11} & \textbf{6.78} & 92.72 & \textbf{59.68} & \textbf{82.35} & \textbf{90.56} \\
BANet-Full & 35.64 & 33 & 37.17 & 210.97 & \textbf{38.50} & 4.20 & 6.94 & \textbf{92.26} & 58.66 & 81.53 & 90.33 \\ \specialrule{1.2pt}{1pt}{1pt}
\end{tabular} } %
\vspace{-2.5ex}
\end{table*}

\begin{table*}[]
\centering
\caption{Quantitative results on the \textbf{DIODE All} (Indoor+Outdoor) \textit{val} set.}
\vspace{-2ex}
\label{tab:diode-all}
\adjustbox{max width=0.85\textwidth}{ %
%\hskip-4.0cm
\begin{tabular}{l|c|c|cccccc|ccc}
\specialrule{1.2pt}{1pt}{1pt}
\multicolumn{1}{c|}{\multirow{2}{*}{Model}} & \multirow{2}{*}{\begin{tabular}[c]{@{}c@{}}\#Params\\ ($10^6$)\end{tabular}} & \multirow{2}{*}{\begin{tabular}[c]{@{}l@{}}Time\\ (ms)\end{tabular}} & \multicolumn{6}{c|}{Lower is better} & \multicolumn{3}{c}{Higher is better (\%)} \\ \cline{4-12}
 &  &  & \textbf{SiLog} & \textbf{SqRel} & \textbf{AbsRel} & \textbf{MAE} & \textbf{RMSE} & \textbf{iRMSE} & $\delta_{25}$ & $\delta_{56}$ & $\delta_{95}$ \\ 
 \specialrule{1.2pt}{1pt}{1pt}

DORN \cite{dorn} & 91.31 & 57 & 34.23 & 226.87 & 47.30 & 3.55 & 5.34 & 135.86 & 46.41 & 73.54 & 86.02 \\
%DORN-BB \cite{dorn} & 91.01 & 50 &  &  &  &  &  &  &  &  &  &  &  &  \\
DenseDepth \cite{densedepth} & 44.61 & 27 & 31.05 & 157.70 & 39.33 & 3.10 & 4.77 & 182.26 & 54.25 & 76.46 & 87.24 \\
BTS \cite{bts} & 47.00 & 34 & 29.83 & \textbf{146.12} & \textbf{37.03} & 3.01 & 4.71 & 152.42 & \textbf{56.35} & 77.29 & \textbf{88.14} \\
\hline
BANet-Vanilla & 28.73 & 21 & 30.11 & 156.16 & 39.30 & 3.08 & 4.73 & 129.02 & 53.76 & 76.67 & 87.37 \\
BANet-Forward & 32.58 & 28 & 30.54 & 181.53 & 39.09 & 3.05 & 4.77 & 128.88 & 55.17 & 76.86 & 87.21 \\
BANet-Backward & 32.58 & 28 & 30.28 & 179.41 & 40.25 & 3.05 & 4.73 & 129.98 & 54.17 & 76.83 & 87.47 \\
BANet-Markov & 35.64 & 31 & 30.16 & 188.45 & 40.10 & 3.12 & 4.80 & 129.31 & 54.04 & 76.04 & 86.76 \\
BANet-Local & 35.08 & 32 & \textbf{29.69} & 148.91 & 37.68 & \textbf{2.98} & \textbf{4.66} & \textbf{126.96} & 56.08 & \textbf{77.34} & 87.96 \\
BANet-Full & 35.64 & 33 & 29.93 & 150.76 & 39.28 & 3.06 & 4.73 & 130.39 & 53.26 & 76.40 & 87.15 \\ \specialrule{1.2pt}{1pt}{1pt}
\end{tabular} } %
\vspace{-2.1ex}
\end{table*}

\begin{table*}[]
\centering
\caption{Quantitative results on the \textbf{KITTI} \textit{val} set.}
\vspace{-2ex}
\label{tab:kitti-val}
\adjustbox{max width=0.85\textwidth}{ %
%\hskip-4.0cm
\begin{tabular}{l|c|c|cccccc|ccc}
\specialrule{1.2pt}{1pt}{1pt}
\multicolumn{1}{c|}{\multirow{2}{*}{Model}} & \multirow{2}{*}{\begin{tabular}[c]{@{}c@{}}\#Params\\ ($10^6$)\end{tabular}} & \multirow{2}{*}{\begin{tabular}[c]{@{}l@{}}Time\\ (ms)\end{tabular}} & \multicolumn{6}{c|}{Lower is better} & \multicolumn{3}{c}{Higher is better (\%)} \\ \cline{4-12}
 &  &  & \textbf{SiLog} & \textbf{SqRel} & \textbf{AbsRel} & \textbf{MAE} & \textbf{RMSE} & \textbf{iRMSE} & $\delta_{25}$ & $\delta_{56}$ & $\delta_{95}$ \\ 
 \specialrule{1.2pt}{1pt}{1pt}

DORN \cite{dorn} & 89.23 & 36 & 12.32 & 2.85 & 11.55 & 1.99 & 3.69 & 11.71 & 91.60 & 98.47 & 99.55 \\
%DORN-BB \cite{dorn} & 91.01 & 50 &  &  &  &  &  &  &  &  &  &  &  &  \\
DenseDepth \cite{densedepth} & 44.61 & 25 & 10.66 & 1.76 & 8.01 & \textbf{1.58} & 3.31 & 8.24 & 93.50 & \textbf{98.87} & 99.69 \\
BTS \cite{bts} & 47.00 & 29 & 10.67 & 1.63 & \textbf{7.64} & \textbf{1.58} & 3.36 & \textbf{8.14} & 93.63 & 98.79 & 99.67 \\
\hline
BANet-Vanilla & 28.73 & 20 & 10.88 & 1.65 & 7.74 & 1.63 & 3.51 & 8.34 & 93.49 & 98.63 & 99.59 \\
BANet-Forward & 32.34 & 25 & 10.61 & 1.67 & 9.02 & 1.77 & 3.54 & 9.99 & 92.66 & 98.68 & 99.61 \\
BANet-Backward & 32.34 & 25 & 10.54 & \textbf{1.52} & 7.67 & 1.59 & 3.42 & 8.47 & 93.57 & 98.72 & 99.65 \\
BANet-Markov & 35.28 & 27 & 10.72 & 1.85 & 8.24 & 1.62 & 3.38 & 8.28 & 93.19 & 98.75 & 99.69 \\
BANet-Local & 35.08 & 27 & \textbf{10.53} & 2.21 & 9.92 & 1.77 & 3.34 & 9.68 & 92.36 & \textbf{98.87} & \textbf{99.71} \\
BANet-Full & 35.28 & 28 & 10.64 & 1.81 & 8.25 & 1.60 & \textbf{3.30} & 8.47 & \textbf{93.75} & 98.81 & 99.68 \\ \specialrule{1.2pt}{1pt}{1pt}
\end{tabular} } %
\vspace{-2.5ex}
\end{table*}

%\vspace{-1ex}
\begin{table}[h!]
\centering
\caption{Results on the KITTI leaderboard (test set).}
\vspace{-2ex}
\label{tab:kitti-leaderboard}
\adjustbox{max width=\textwidth}{
\begin{tabular}{l|cccc}
\specialrule{1.2pt}{1pt}{1pt}
\multicolumn{1}{c|}{Method} & SILog & SqRel & AbsRel & iRMSE \\ \specialrule{1.2pt}{1pt}{1pt}
DORN \cite{dorn} & 11.77 & 2.23 & \textbf{8.78} & 12.98 \\
BTS \cite{bts} & 11.67 & \textbf{2.21} & 9.04 & 12.23 \\
\textbf{BANet-Full} & \textbf{11.55} & 2.31 & 9.34 & \textbf{12.17} \\ \specialrule{1.2pt}{1pt}{1pt}
\end{tabular} } %
\vspace{-0.7cm}
\end{table}

%%%%%%%%%%%%%%%%%%%%%%%%%%%%%%%%%%%
\subsection{Results on Monocular Depth Estimation}
\noindent \textbf{Quantitative Comparison:} Table \ref{tab:diode-indoor}, \ref{tab:diode-outdoor}, \ref{tab:diode-all}, and \ref{tab:kitti-val}  present the comparison of our BANet variants with SOTA architectures. Note that DORN performs much worse due to the effect of granularity provided by its discretization or ordinal levels. A higher value of this parameter might improve its precision, but with exponential increase in memory consumption during training; thus making it difficult to train on large datasets like KITTI and DIODE. Overall, our BANet variants perform better or close to existing SOTA in particular BTS. 
%It is worth noting that the number of parameters in the heaviest BANet variant (BANet-Full) is about 25\% less than BTS. 
%The number of parameters in BANet varies slightly with the input resolution due to the variation in the number of parameters in FC layer after average pooling with the predefined kernel size ($32\times 32$). 
Finally, our model performs better than SOTA architectures on the KITTI leaderboard \footnote[1]{\url{http://www.cvlibs.net/datasets/kitti/eval_depth.php?benchmark=depth_prediction}} (Table~\ref{tab:kitti-leaderboard}) in terms of the primary metric (SILog) used for ranking.

%%%%%%%%%%%%%%%%%%%%%%%%%%%%%%
\begin{figure*}[t]
\centering
\adjustbox{max width=0.95\textwidth}{
\includegraphics[width=\textwidth]{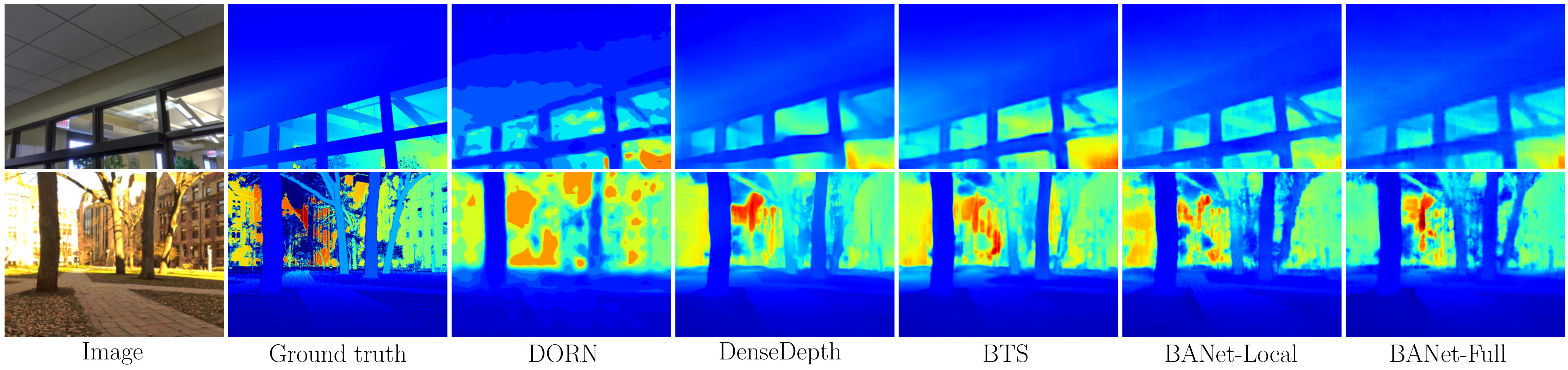}
} %
\vspace{-0.2cm}
\caption{Sample results on DIODE indoor (top) and outdoor (bottom) images. \textbf{Top:} The window frame on the bottom right is detected well by BANets compared to BTS and DenseDepth. \textbf{Bottom:} Other methods but our BANet-Full fail to localize the highly illuminated tree trunks. This shows the importance of global context aggregation in BANet-Full.}
\label{fig:diode_results}
\vspace{-0.5cm}
\end{figure*}
%%%%%%%%%%%%%%%%%

\begin{figure}[t!]
\centering
\adjustbox{max width=0.45\textwidth}{
\includegraphics[width=0.4\textheight]{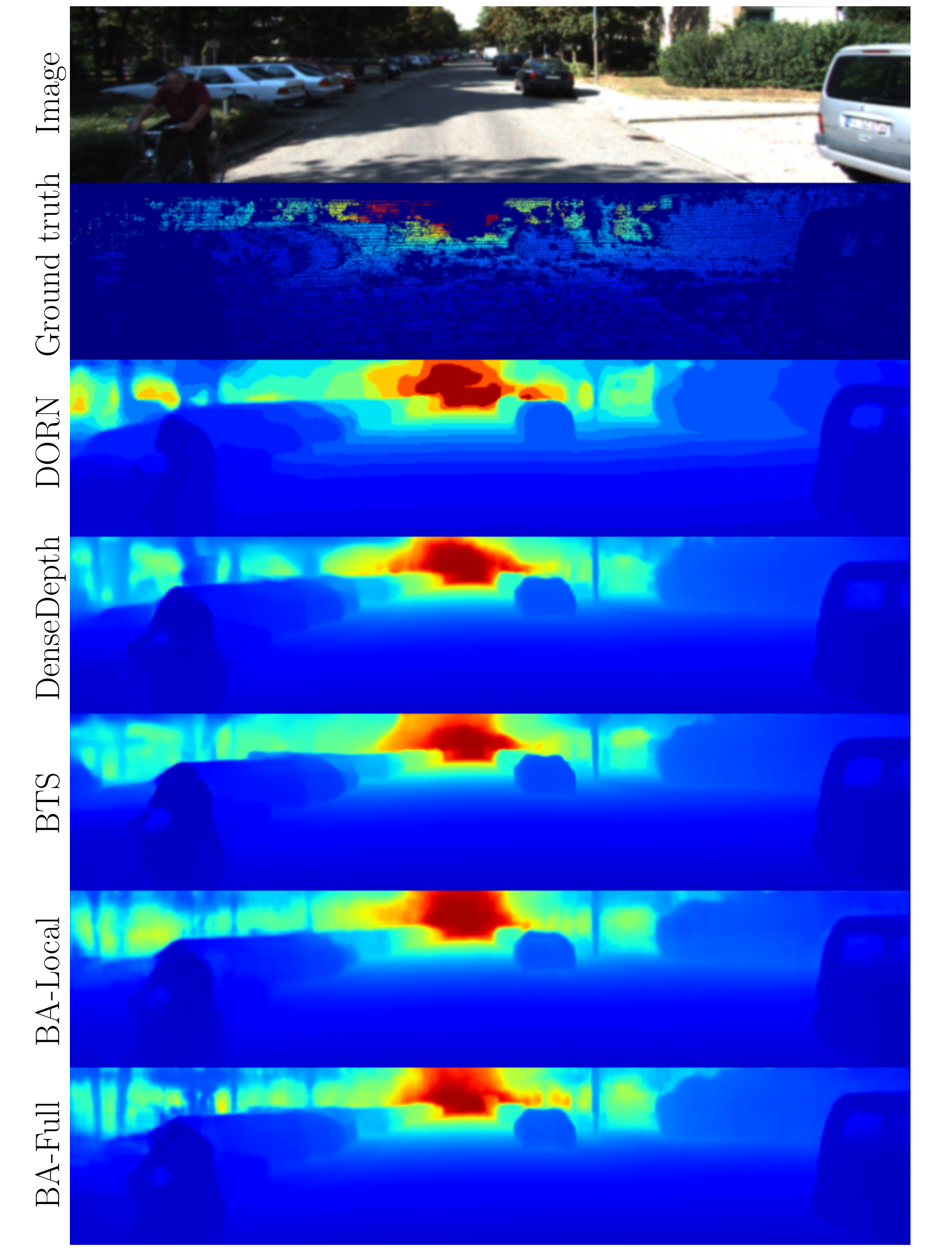}
} %
\vspace{-0.2cm}
\caption{Sample results on KITTI \textit{val} set. The top region is cropped since it does not contain any ground truth. The tree trunks in the dark regions on the left is better differentiated with global context aggregation in BANet-Full.}
\label{fig:kitti_results}
\vspace{-0.5cm}
\end{figure}

\vspace{1ex}
\noindent \textbf{Qualitative Comparison:} Figure \ref{fig:diode_results} and \ref{fig:kitti_results} show the qualitative comparison of different SOTA methods. Also, we provide a video of our predictions for the KITTI validation sequences \footnote[2]{\url{https://www.youtube.com/watch?v=a9wW02pfeKw&ab_channel=MannatKaur}}. The DORN prediction map clearly indicates its semi-discretized nature of prediction. For both the indoor and outdoor images in Fig.~\ref{fig:diode_results}, BTS and DenseDepth suffer from the high intensity regions surrounding the window frame in the bottom right of the left image and the trunks in the right image. Our BANet variants work better in these delicate regions. 
%In particular, the global context aggregation of BANet-Full helps to distinguish the trunks better than its local counterpart. Fig.~\ref{fig:kitti_results} demonstrates a similar case, but this time in the dark surroundings. From both these visualizations, it is evident that the global context aggregation is necessary to resolve potential ambiguities caused by adverse illumination.
To better tease out the effect of the different components in BANet, we provide observations as follows: 
\vspace{1ex}\\
\noindent $\bullet$ \textbf{Why does vanilla perform better than SOTA in many cases?} The vanilla architecture inherently maintains and generates the full-resolution depth map using the D2S conversion without increasing the number of parameters. We hypothesize this to be the primary reason why it performs better than DORN (huge parameters) and DenseDepth (generates $4\times$ downsampled output followed by $4\times$ upsampling).
\vspace{1ex}\\
\noindent $\bullet$ \textbf{Why is Markovian worse than vanilla?} BANet-Markov utilizes only the $S_{i - 1}$ stage features to generate the attention map for $S_i$ stage features. 
%Note that the backbone already computes the $S_i$ stage features taking $S_{i-1}$ as input. 
%Therefore, for BANet-Markov, $s_i = \mathcal{F}(s_{i-1}) * \mathcal{A}(s_{i-1}) $ where $\mathcal{F}$ and $\mathcal{A}$ are the composite nonlinear functions representing the feature and attention modules operating on the exact same input $s_{i-1}$ unlike other variants where the attentions for one stage are computed from all the previous (forward mode) or later (backward mode) stages inclusive. In other words, 
For other variants, when $s_{i-1}, s_{i-2}, s_{i-3}, \hdots$ are used for the attention mechanism, the forgotten (or future for backward) information from $s_{i-2}, s_{i-3}, \hdots$ are used for the refinement. However, essentially no forgotten information is brought back to the stage for Markovian formulation. The same information (or $s_{i-1}$) that was directly used to compute the feature map $s_i$ is also being reused for attention. We believe this lack of forgotten information prevent the BANet-Markov variant to gain a clear edge compared to the vanilla architecture. 
\vspace{1ex}\\
\noindent $\bullet$ \textbf{Backward refinement is more crucial than the forward:} The superior performance of BANet-Backward compared to BANet-Forward indicates that the feature refinement is potentially more important in the early stages of the backbone than later in terms of performance.
%\vspace{1ex}\\
%\noindent \textbf{Improved performance with time-dependent structuring improves performance with negligible overhead:} The time-dependent structure inscribed into the model improves the performance over the vanilla counterparts with little increase in computational and memory complexity.
\vspace{1ex}\\
\noindent $\bullet$ \textbf{Effect of global context aggregation:} The effect of global context aggregation cannot be well-understood with the numerical differences in performance. This is because it mostly attempts to refine the prediction for the delicate regions like the thin tree trunks and branches both in extremely dark or bright backgrounds (Figure \ref{fig:diode_results} and \ref{fig:kitti_results}). Note that the pixel coverage of such critical regions is quite low compared to the total number of annotated pixels, and the vanilla and local variants perform well for these (comparatively easier) regions. However, in a real scenario, it is more important to identify the related depth of both the difficult and easy pixels satisfactorily than to be better only at the easy pixels, which are in bulk quantities in all the datasets. In other words, compared to the other variants, BANet-Full has a slightly higher residual error with much less number of false negatives which is critical for sensitive applications like autonomous driving. Therefore, based upon our understanding, BANet-Full is a more reliable alternative than the others in almost all the cases even though it is slightly off in numbers. 
\vspace{1ex}\\
\noindent $\bullet$ \textbf{General pattern of stage-wise attention:} The backward attention dominates the forward ones (Figure \ref{fig:pseudo}) because the raw, stage-wise feature maps are computed with the backbone in the forward direction, thus encoding more forward information. The general pattern of stage-wise bidirectional attention can be described as follows:\\
\noindent \textbf{Stage-1:} Generic activations spread over the whole scene.\\
\noindent \textbf{Stage-2:} Emphasis is on the ground plane or road regions with gradual change in depth from the observer.\\
\noindent \textbf{Stage-3:} Focus is on the individual objects, such as cars, pedestrians, etc. This one assists in measuring the ground plane distance of the objects from the observer's location.\\
\noindent \textbf{Stage-4:} Focus is still on the objects with a gradual change within each object from the ground towards the top. Therefore, this is measuring the intra-object depth variation.\\
\noindent \textbf{Stage-5:} Primarily devoted to the understanding of the vanishing point information to get a sense of the upper bound of depth in the scene in general.
%\end{itemize} 

\section{Conclusion}

In this paper, we present a bidirectional attention mechanism for convolutional networks for monocular depth estimation task similar to the bidirectional RNN for NMT. We show that a standard backbone equipped with this attention mechanism performs better or at least on a par with the SOTA architectures on the standard datasets. As a future direction, we believe that our proposed attention mechanism has the potential to be useful in similar dense prediction tasks, esp. the single plane ones.  

\bibliographystyle{IEEEtran}
\bibliography{perception,mde,attention,arch}

\begin{thebibliography}{10}
\providecommand{\url}[1]{#1}
\csname url@samestyle\endcsname
\providecommand{\newblock}{\relax}
\providecommand{\bibinfo}[2]{#2}
\providecommand{\BIBentrySTDinterwordspacing}{\spaceskip=0pt\relax}
\providecommand{\BIBentryALTinterwordstretchfactor}{4}
\providecommand{\BIBentryALTinterwordspacing}{\spaceskip=\fontdimen2\font plus
\BIBentryALTinterwordstretchfactor\fontdimen3\font minus
  \fontdimen4\font\relax}
\providecommand{\BIBforeignlanguage}[2]{{%
\expandafter\ifx\csname l@#1\endcsname\relax
\typeout{** WARNING: IEEEtran.bst: No hyphenation pattern has been}%
\typeout{** loaded for the language `#1'. Using the pattern for}%
\typeout{** the default language instead.}%
\else
\language=\csname l@#1\endcsname
\fi
#2}}
\providecommand{\BIBdecl}{\relax}
\BIBdecl

\bibitem{sensation-and-perception-goldstein}
E.~Goldstein, \emph{Sensation and Perception}.\hskip 1em plus 0.5em minus
  0.4em\relax Cengage Learning, 2009.

\bibitem{taxonomy-szeliski}
D.~Scharstein and R.~Szeliski, ``A taxonomy and evaluation of dense two-frame
  stereo correspondence algorithms,'' \emph{IJCV}, vol.~47, 2002.

\bibitem{compare-szeliski}
S.~M. {Seitz}, B.~{Curless}, J.~{Diebel}, D.~{Scharstein}, and R.~{Szeliski},
  ``A comparison and evaluation of multi-view stereo reconstruction
  algorithms,'' in \emph{CVPR}, 2006.

\bibitem{densedepth}
I.~Alhashim and P.~Wonka, ``High quality monocular depth estimation via
  transfer learning,'' \emph{CoRR}, vol. abs/1812.11941, 2018.

\bibitem{bts}
J.~H. Lee, M.~Han, D.~W. Ko, and I.~H. Suh, ``From big to small: Multi-scale
  local planar guidance for monocular depth estimation,'' \emph{CoRR}, vol.
  abs/1907.10326, 2019.

\bibitem{mde-as-classification}
Y.~{Cao}, Z.~{Wu}, and C.~{Shen}, ``Estimating depth from monocular images as
  classification using deep fully convolutional residual networks,'' \emph{IEEE
  Transactions on Circuits and Systems for Video Technology}, vol.~28, no.~11,
  2018.

\bibitem{dorn}
H.~Fu, M.~Gong, C.~Wang, K.~Batmanghelich, and D.~Tao, ``Deep ordinal
  regression network for monocular depth estimation,'' in \emph{CVPR}, 2018.

\bibitem{vgg}
K.~Simonyan and A.~Zisserman, ``Very deep convolutional networks for
  large-scale image recognition,'' in \emph{ICLR}, 2015.

\bibitem{resnet}
K.~{He}, X.~{Zhang}, S.~{Ren}, and J.~{Sun}, ``Deep residual learning for image
  recognition,'' in \emph{CVPR}, 2016.

\bibitem{densenet}
G.~{Huang}, Z.~{Liu}, L.~{Van Der Maaten}, and K.~Q. {Weinberger}, ``Densely
  connected convolutional networks,'' in \emph{CVPR}, 2017.

\bibitem{amirul2017gated}
M.~Amirul~Islam, M.~Rochan, N.~D. Bruce, and Y.~Wang, ``Gated feedback
  refinement network for dense image labeling,'' in \emph{CVPR}, 2017.

\bibitem{d2s}
W.~Shi, J.~Caballero, F.~Huszar, J.~Totz, A.~P. Aitken, R.~Bishop, D.~Rueckert,
  and Z.~Wang, ``Real-time single image and video super-resolution using an
  efficient sub-pixel convolutional neural network,'' in \emph{CVPR}, 2016.

\bibitem{d2s-aich}
S.~{Aich}, W.~{van der Kamp}, and I.~Stavness, ``Semantic binary segmentation
  using convolutional networks without decoders,'' in \emph{CVPRW}, 2018.

\bibitem{how-see-depth}
T.~v. Dijk and G.~d. Croon, ``How do neural networks see depth in single
  images?'' in \emph{ICCV}, 2019.

\bibitem{visualize-mde}
J.~Hu, Y.~Zhang, and T.~Okatani, ``Visualization of convolutional neural
  networks for monocular depth estimation,'' in \emph{ICCV}, 2019.

\bibitem{nmt-attention}
D.~Bahdanau, K.~Cho, and Y.~Bengio, ``Neural machine translation by jointly
  learning to align and translate,'' in \emph{ICLR}, 2015.

\bibitem{senet}
J.~Hu, L.~Shen, and G.~Sun, ``Squeeze-and-excitation networks,'' in
  \emph{CVPR}, 2018.

\bibitem{cbam}
S.~Woo, J.~Park, J.-Y. Lee, and I.~S. Kweon, ``Cbam: Convolutional block
  attention module,'' in \emph{ECCV}, 2018.

\bibitem{encnet}
H.~Zhang, K.~Dana, J.~Shi, Z.~Zhang, X.~Wang, A.~Tyagi, and A.~Agrawal,
  ``Context encoding for semantic segmentation,'' in \emph{CVPR}, 2018.

\bibitem{ecanet}
Q.~Wang, B.~Wu, P.~Zhu, P.~Li, W.~Zuo, and Q.~Hu, ``Eca-net: Efficient channel
  attention for deep convolutional neural networks,'' in \emph{CVPR}, 2020.

\bibitem{attention-aug}
I.~Bello, B.~Zoph, A.~Vaswani, J.~Shlens, and Q.~V. Le, ``Attention augmented
  convolutional networks,'' in \emph{ICCV}, 2019.

\bibitem{saxena-2006}
A.~Saxena, S.~H. Chung, and A.~Y. Ng, ``Learning depth from single monocular
  images,'' in \emph{NIPS}, 2006.

\bibitem{make3d-v1}
A.~{Saxena}, S.~H. {Chung}, and A.~Y. {Ng}, ``Learning depth from single
  monocular images,'' in \emph{NIPS}, 2006.

\bibitem{make3d-v3}
A.~{Saxena}, M.~{Sun}, and A.~Y. {Ng}, ``Make3d: Learning 3d scene structure
  from a single still image,'' \emph{IEEE Transactions on Pattern Analysis and
  Machine Intelligence}, vol.~31, no.~5, pp. 824--840, 2009.

\bibitem{eigen}
D.~Eigen, C.~Puhrsch, and R.~Fergus, ``Depth map prediction from a single image
  using a multi-scale deep network,'' in \emph{NIPS}, 2014.

\bibitem{li-crf}
B.~{Li}, C.~{Shen}, Y.~{Dai}, A.~{van den Hengel}, and M.~{He}, ``Depth and
  surface normal estimation from monocular images using regression on deep
  features and hierarchical crfs,'' in \emph{CVPR}, 2015.

\bibitem{dcnf}
F.~{Liu}, C.~{Shen}, G.~{Lin}, and I.~{Reid}, ``Learning depth from single
  monocular images using deep convolutional neural fields,'' \emph{IEEE TPAMI},
  vol.~38, no.~10, 2016.

\bibitem{ms-crf}
D.~{Xu}, E.~{Ricci}, W.~{Ouyang}, X.~{Wang}, and N.~{Sebe}, ``Multi-scale
  continuous crfs as sequential deep networks for monocular depth estimation,''
  in \emph{CVPR}, 2017.

\bibitem{structured-ms-crf}
D.~Xu, W.~Wang, H.~Tang, H.~Liu, N.~Sebe, and E.~Ricci, ``Structured attention
  guided convolutional neural fields for monocular depth estimation,'' in
  \emph{CVPR}, 2018.

\bibitem{alexnet}
A.~Krizhevsky, I.~Sutskever, and G.~E. Hinton, ``Imagenet classification with
  deep convolutional neural networks,'' in \emph{NIPS}, 2012.

\bibitem{inception-v3}
C.~Szegedy, V.~Vanhoucke, S.~Ioffe, J.~Shlens, and Z.~Wojna, ``Rethinking the
  inception architecture for computer vision,'' in \emph{CVPR}, 2016.

\bibitem{refinedmpl}
J.~M.~U. Vianney, S.~Aich, and B.~Liu, ``Refinedmpl: Refined monocular
  pseudolidar for 3d object detection in autonomous driving,'' 2019.

\bibitem{nyuv2}
N.~{Silberman}, D.~{Hoiem}, P.~{Kohli}, and R.~{Fergus}, ``Indoor segmentation
  and support inference from rgbd images,'' in \emph{ECCV}, 2012.

\bibitem{diode}
I.~Vasiljevic, N.~Kolkin, S.~Zhang, R.~Luo, H.~Wang, F.~Z. Dai, A.~F. Daniele,
  M.~Mostajabi, S.~Basart, M.~R. Walter, and G.~Shakhnarovich, ``{DIODE}: {A}
  {D}ense {I}ndoor and {O}utdoor {DE}pth {D}ataset,'' \emph{CoRR}, vol.
  abs/1908.00463, 2019.

\bibitem{kitti}
J.~{Uhrig}, N.~{Schneider}, L.~{Schneider}, U.~{Franke}, T.~{Brox}, and
  A.~{Geiger}, ``Sparsity invariant cnns,'' in \emph{3DV}, 2017.

\end{thebibliography}

\end{document}